\documentclass[10pt,twocolumn,letterpaper]{article}

\usepackage[algorithms]{wacv}

\usepackage{graphicx}
\usepackage{subcaption}
\usepackage{enumitem}
\usepackage{amsmath}
\usepackage{amssymb}
\usepackage{booktabs}
\usepackage{colortbl}
\definecolor{iilgray}{gray}{0.88}
\usepackage{multirow}
\usepackage{makecell}
\usepackage{array}
\usepackage{placeins}
\usepackage{microtype}

\usepackage{algorithm}
\usepackage{algpseudocode}

\graphicspath{{figures/}}

\newcommand{\Lcls}{\mathcal{L}_{\text{cls}}}
\newcommand{\LIRM}{\mathcal{L}_{\text{IRMv1}}}
\newcommand{\LREx}{\mathcal{L}_{\text{REx}}}
\newcommand{\bx}{\mathbf{x}}
\newcommand{\E}{\mathbb{E}}
\newcommand{\R}{\mathbb{R}}

\renewcommand{\eg}{\textit{e.g.}}
\renewcommand{\etal}{\textit{et al.}}

\newcommand{\mypar}[1]{\smallskip\noindent\textbf{#1.}}

\def\wacvPaperID{}
\def\confName{WACV}
\def\confYear{2027}

\definecolor{wacvblue}{rgb}{0.21,0.49,0.74}
\usepackage[breaklinks,colorlinks,allcolors=wacvblue]{hyperref}

\title{Intersectional Invariant Learning for Fair Deepfake Detection}

\author{Uzoamaka Ezeakunne$^{1}$\quad Taiwo Olabamipe$^{2}$\quad Deevesh Jain$^{1}$\quad Xiuwen Liu$^{1}$\\
$^{1}$Department of Computer Science, Florida State University\\
$^{2}$Department of Computer Science, University of Central Florida\\
{\tt\small uezeakunne@fsu.edu, ta117847@ucf.edu, dj23f@fsu.edu, xliu@fsu.edu}
}

\begin{document}
\maketitle

\begin{abstract}
Deepfake detection models achieve high overall accuracy but exhibit large
performance disparities across demographic groups, causing some groups to
experience substantially higher false-positive rates than others.
Existing fairness-aware methods address this by constraining model outputs to
equalize group predictions, but do not tackle the underlying feature-level bias.
As a result, models may still rely on demographic characteristics, such as skin
texture, as indirect cues for detecting manipulated content.
To address this problem, we propose
\emph{Intersectional Invariant Learning} (IIL), a framework that
improves feature learning rather than only adjusting model outputs.
IIL alternates between identifying the most challenging demographic partitions
and learning features that remain consistent across those partitions.
The framework also includes a false-positive-rate fairness objective and
group-balanced sampling across six gender${\times}$skin-tone intersectional groups.
We evaluate IIL on a large-scale fairness benchmark containing over one million
images, 10 baseline detectors, and 21 fairness and utility metrics.
Results show that IIL achieves Pareto-optimal performance, simultaneously
attaining an AUC above 99\%, a false-positive rate under 2.5\%, and the
lowest intersectional fairness disparity (FMEO) among all compared methods.
IIL outperforms prior fairness-aware methods on the large majority of the
21 evaluation metrics, while reducing both false-positive rates and
intersectional fairness disparities by several-fold compared with the
strongest competing method.
\end{abstract}

\section{Introduction}
\label{sec:intro}

Deepfake detectors usually report an average AUC greater than 98\%, yet
such aggregate metrics mask systematic disparities: on the AI-Face
benchmark~\cite{aiface}, EfficientNet-B4~\cite{tan2019efficientnet} reaches
a Skin-Tone FMEO of 8.3\%, and even explicitly fairness-aware detectors
(DAG-FDD~\cite{ju2024improving}, PG-FDD~\cite{lin2024pgfdd}) leave
intersectional FMEO at 12.1\% and 9.5\% respectively, each more than
$6{\times}$ our result of 1.46\%. Prior work has identified demographic
subgroups with large error-rate disparities~\cite{trinh2021fairness,xu2022bias},
with real-world impact when deployed in content moderation or legal
proceedings~\cite{gong2025deepfakeimpact,wang2024survey}.

\mypar{Why demographic parity regularization is insufficient}
A common approach penalizes the prediction-rate gap across
demographic groups (FDP)~\cite{agarwal2018reductions}, but this is
inadequate: FDP ignores False Positive Rate and acts on the model's
\emph{output}, not its \emph{representation}.
GradCAM~\cite{selvaraju2017gradcam} (Figure~\ref{fig:attention}) shows
EffB4/DAG activate broadly over facial skin on real images regardless of
demographic group --- skin texture acts as a spurious cue rather than a
landmark-anchored one, a feature-level pathology output penalization
cannot address.

\mypar{Invariant learning for fairness}
Invariant Risk Minimization (IRM)~\cite{arjovsky2019irm} and Risk
Extrapolation (REx)~\cite{krueger2021rex} learn representations predictive
\emph{across} environments, directly blocking the feature-level
mechanisms that produce FMEO and FEO gaps.
Ma~\etal~\cite{ma2023ifr} apply this to face recognition; concurrent
work~\cite{cheng2025fair} shows fair detectors can retain generalization but
does not address FPR disparity's feature-level origins.
To our knowledge, IIL is the first application of invariant learning to
deepfake detection, with three detection-specific additions: an explicit
\textbf{FPR parity loss} targeting the gap FDP misses;
\textbf{group-balanced sampling} to prevent dominant groups from
overwhelming rare ones; and \textbf{$K{=}6$ intersectional partitions}, one
per gender${\times}$skin-tone group.

IIL is Pareto-dominant on the AI-Face benchmark, achieving the highest AUC
(99.42\%) and the lowest intersectional FMEO (1.46\%) simultaneously,
leading on 17 of 21 metrics with $4.6{\times}$ lower FPR and $6.5{\times}$
lower inter-FMEO than the best competing fairness-aware detector
(PG-FDD~\cite{lin2024pgfdd}).
GradCAM analysis (Section~\ref{sec:analysis}) ties these gains to \emph{where}
the model attends: broad, skin-spanning activation on real images is
replaced by attention anchored to specific facial structures, distinct for
real versus fake images and reproducible across independently resampled
images --- a targeted attention change, not a general removal of
demographic information from the representation (quantified in
Supplementary Section~\ref{sec:supp_invariance_probe}), consistent with IIL's risk-invariance mechanism
(Section~\ref{sec:method}).

\section{Related Work}
\label{sec:related}

\mypar{Deepfake detection}
FaceForensics++~\cite{rossler2019faceforensics} established Xception as the
dominant detection baseline. Later work focused on frequency-based
artifacts (F3Net~\cite{li2021f3net}, SPSL~\cite{liu2021spsl},
SRM~\cite{luo2021srm}), generalization across unseen forgeries
(UCF~\cite{yan2023ucf}, CORE~\cite{ni2022core}, LSDA~\cite{yan2024lsda},
Effort~\cite{yan2025effort}), and fairness-aware detection
(DAW-FDD~\cite{ju2024improving}, DAG-FDD~\cite{ju2024improving},
PG-FDD~\cite{lin2024pgfdd}). Several benchmarks and
surveys~\cite{gong2025deepfakeimpact,wang2024survey,pei2024survey} document
that strong overall accuracy can hide large, skin-tone- and
gender-correlated error-rate gaps~\cite{trinh2021fairness,xu2022bias}, and
that demographic-parity constraints alone are not sufficient: the AI-Face
benchmark~\cite{aiface} shows fairness-trained models can still exhibit
substantial demographic disparities.
Recent fairness-aware detectors pursue this goal through different
mechanisms: Ding~\etal~\cite{ding2026decoupling} decouple
fairness-sensitive channels and align group-wise prediction distributions
with optimal transport; FairReL~\cite{lu2026fairrel} learns fairness-aware
representations; and FairForensics~\cite{zhang2026fairforensics} uses
vision-language supervision with demographic prompts. Earlier data-driven
work combined demographically balanced synthetic augmentation with
multi-task demographic learning to improve fairness generalization
across datasets~\cite{ezeakunne2025icaart}, and post-hoc calibration on
frozen face embeddings has also been proposed~\cite{brown2026fft}.

\mypar{Group fairness and invariant learning}
Demographic Parity~\cite{dwork2012fairness} requires equal prediction rates
across groups (FDP); Equalized Odds~\cite{hardt2016equality} additionally
requires equal TPR and FPR, a stronger criterion more suitable for this
work. GroupDRO~\cite{sagawa2019dro} improves worst-group performance but
does not explicitly learn shared representations across groups.
IRM~\cite{arjovsky2019irm} and REx~\cite{krueger2021rex} learn features
stable across environments; Ma~\etal~\cite{ma2023ifr} applied IRM to face
recognition. Yin~\etal~\cite{yin2024id3} used invariant risk
minimization to improve the cross-domain generalization of deepfake
detectors, without addressing demographic fairness. Building on these ideas, and unlike the fairness-aware detectors above,
IIL applies IRMv1 and REx penalties across the most unfavorable
demographic partitions, combining fairness optimization
and invariant representation learning for deepfake detection, targeting
false positives caused by demographic-specific facial characteristics
such as skin texture.

\section{Problem Setting}
\label{sec:problem}

Let $\mathcal{D} = \{(\bx_i, y_i, g_i)\}_{i=1}^{N}$ where
$\bx_i \in \R^{H\times W\times 3}$ is a face image,
$y_i \in \{0,1\}$ is the real/fake label, and
$g_i \in \{0,\dots,5\}$ indexes a demographic intersection group.
The AI-Face dataset~\cite{aiface} provides three demographic axes:
\textbf{skin tone} (Fitzpatrick scales 1--3, 4--6, and 7--10 collapsed
to Light, Medium, Dark), \textbf{gender} (Male / Female), and
\textbf{predicted age} (Child / Young / Adult / Middle / Senior).
The gender${\times}$skin-tone combinations yield $G{=}6$ intersectional
groups: Female/Male $\times$ Light/Medium/Dark.
Group sizes are severely imbalanced: Female-Medium (${\approx}485$k images
in the training split) is over $105{\times}$ larger than the smallest
group, Male-Light (${\approx}4.6$k), a significant source of
minority-group under-representation under uniform sampling.

A detector $f_\theta : \bx \mapsto [0,1]$ predicts the probability of an
image being a deepfake. Let $\hat{y}_i = \mathbf{1}[f_\theta(\bx_i) > 0.5]$,
and let $\text{TPR}_g$, $\text{FPR}_g$, $\text{Acc}_g$ denote per-group rates.

\subsection{Metrics}
\label{sec:metrics}

We follow the evaluation protocol of~\cite{aiface},
computing all metrics on the standard test split
(10\,291 batches of size 32, 329{,}309 test images).

\mypar{Utility} (\textit{higher} $\uparrow$ / \textit{lower} $\downarrow$ is better):
\textbf{Acc}$\uparrow$, \textbf{AUC}$\uparrow$, \textbf{AP}$\uparrow$,
\textbf{EER}$\downarrow$, \textbf{FPR}$\downarrow$.

\mypar{Fairness} (all $\downarrow$; lower gap = fairer):
\begin{itemize}[leftmargin=*, nosep]
  \item \textbf{FDP} --- Demographic parity gap:
    $\max_g \E[\hat{y}|g] - \min_g \E[\hat{y}|g]$: how much more often
    one group is flagged fake than another, regardless of ground truth.
  \item \textbf{FMEO} --- Max Equalized Odds gap:
    $\max\!\bigl(\max_g\text{TPR}_g{-}\min_g\text{TPR}_g,\;
    \max_g\text{FPR}_g{-}\min_g\text{FPR}_g\bigr)$: the worst-case gap in
    either catching forgeries or falsely accusing real images.
  \item \textbf{FEO} --- Summed Equalized Odds deviation:
    $\sum_g\!\bigl(|\text{FPR}_g{-}\overline{\text{FPR}}|
    + |\text{TPR}_g{-}\overline{\text{TPR}}|\bigr)$: like FMEO but summed
    over all groups, so it also penalizes disparities spread broadly
    rather than concentrated in one outlier.
  \item \textbf{FOAE} --- Overall accuracy equity:
    $\max_g\text{Acc}_g - \min_g\text{Acc}_g$: how much less accurate the
    detector is for one group overall, regardless of error type.
\end{itemize}

FMEO is our primary fairness target: it captures the worst-case parity gap
in \emph{both} detection rate and false-positive rate, quantifying the
asymmetric impact described in Section~\ref{sec:intro}. FDP is secondary
--- it is achievable without reducing per-group FPR
disparities~\cite{hardt2016equality}, making it an insufficient sole
objective.

\section{Intersectional Invariant Learning}
\label{sec:method}

IIL combines invariant risk minimization (IRM~\cite{arjovsky2019irm}) and
risk extrapolation (REx~\cite{krueger2021rex}) with detection-specific
fairness losses for deepfake detection without identity labels
(Table~\ref{tab:design} compares components to prior work). Prior
work~\cite{ma2023ifr} uses a supervised contrastive
loss~\cite{khosla2020supcon} for its many fine-grained per-identity
partitions; IIL's six coarse intersectional groups are handled directly
by the classification head, needing no contrastive objective.

\begin{table}[t]
\centering
\caption{IIL components compared with prior invariant-learning
work~\cite{ma2023ifr}. $\checkmark$\,=\,present; $\times$\,=\,absent.}
\label{tab:design}
\resizebox{\columnwidth}{!}{%
\begin{tabular}{lcc}
\toprule
\textbf{Component} & \textbf{Prior work (INV-REG)~\cite{ma2023ifr}} & \textbf{IIL} \\
\midrule
Step 1: REx~\cite{krueger2021rex}     & $\checkmark$ & $\checkmark$ \\
Step 2: IRMv1~\cite{arjovsky2019irm}  & $\checkmark$ & $\checkmark$ \\
Classification loss $\Lcls$           & Supcon.~\cite{khosla2020supcon} & BCE \\
Partition level                       & Identity      & Gender${\times}$Skin-tone \\
$K$ (partitions)                      & 2             & 6            \\
FPR parity loss                       & $\times$      & $\checkmark$ \\
Group-balanced sampling               & $\times$      & $\checkmark$ \\
Multi-criteria fairness targets       & $\times$      & $\checkmark$ \\
\bottomrule
\end{tabular}}
\end{table}

\subsection{Invariant Loss Objectives}

\mypar{IRMv1 (Step 2)}
Given $K$ partition subsets, IRMv1 penalizes the gradient of a scalar dummy
classifier $w{=}1$ on each subset's outputs:
\begin{equation}
  \LIRM = \sum_{k=1}^{K} L_k +
    \lambda_{\text{IRM}} \sum_{k=1}^{K}
    \bigl\|\nabla_{w|w=1} L_k(w \cdot f)\bigr\|^2,
  \label{eq:irm}
\end{equation}
where $L_k = \text{BCE}(f(\bx^k), y^k)$.
The gradient penalty is zero iff $w{=}1$ is simultaneously optimal on all
subsets --- enforcing that no subset-specific cue (\eg, skin-tone texture) can
improve predictions.
Minimizing this penalty directly reduces the feature-level causes of FMEO and FEO.

\mypar{REx (Step 1)}
Risk Extrapolation penalizes the \emph{variance} of per-partition losses:
\begin{equation}
  \LREx = \sum_{k=1}^{K} L_k +
    \lambda_{\text{REx}} \cdot \text{Var}(\{L_k\}_{k=1}^K).
  \label{eq:rex}
\end{equation}
REx is used only in Step~1 (partition learning) with the detector outputs
\emph{detached}, providing a smoother gradient signal than IRMv1 for
maximizing worst-case partition divergence~\cite{krueger2021rex}.

\subsection{Algorithm}

\begin{algorithm}[!t]
\caption{IIL (Intersectional Invariant Learning)}
\label{alg:inv-reg}
\begin{algorithmic}[1]
\Require Dataset $\mathcal{D}$ sampled with a group-balanced WeightedRandomSampler;
detector $f_\theta$; hyperparameters $\lambda_{\mathrm{IRM}}$, $\lambda_{\mathrm{REx}}$, $\lambda$;
number of partitions $K=6$; partition-learning steps $T$; learning rate $\eta$
\State \textbf{Pre-train} $f_\theta$ for one epoch using $\mathcal{L}_{\mathrm{BCE}}$, $\mathcal{L}_{\mathrm{DP}}$, $\mathcal{L}_{\mathrm{FPR}}$.
\For{each subsequent epoch}
    \For{each mini-batch $(\mathbf{x}, y, g)$ from the balanced sampler}
        \State \textbf{Step 1: Partition Learning} \Comment{freeze $f_\theta$}
        \State Initialize $\mathbf{P} \sim \mathcal{N}(0, 0.01)$, where $\mathbf{P} \in \mathbb{R}^{G \times K}$.
        \Comment{random initialization breaks symmetry}
        \For{$t = 1,\ldots,T$}
            \State Compute $\mathcal{L}_{\mathrm{REx}}$ using detached predictions from $f_\theta(\mathbf{x})$.
            \State $\mathbf{P} \leftarrow \mathbf{P} + \eta \nabla_{\mathbf{P}}\mathcal{L}_{\mathrm{REx}}$
            \Comment{gradient ascent on $\mathcal{L}_{\mathrm{REx}}$}
        \EndFor
        \State $P_g^* \leftarrow \arg\max_k \mathbf{P}_{g,k}$ for each group $g$.
        \State \textbf{Step 2: Feature Learning} \Comment{update $\theta$}
        \State Compute $\mathcal{L}_{\mathrm{IRM}}$ using live predictions from $f_\theta(\mathbf{x})$ under $P^*$.
        \State $\mathcal{L}_{\mathrm{DP}} \leftarrow \max_g r_g - \min_g r_g$
        \Comment{$r_g = \mathrm{mean}(f_\theta(\mathbf{x}) \mid G{=}g)$; prediction-rate parity}
        \State $\mathcal{L}_{\mathrm{FPR}} \leftarrow
        \max_g \widehat{\mathrm{FPR}}_g -
        \min_g \widehat{\mathrm{FPR}}_g$
        \Comment{$y=0$ only; $\widehat{\mathrm{FPR}}_g =
        \mathrm{mean}(f_\theta(\mathbf{x}) \mid y=0, G=g)$}
        \State $\mathcal{L} \leftarrow
        \mathcal{L}_{\mathrm{BCE}}
        + \lambda \left(
        \mathcal{L}_{\mathrm{DP}}
        + \mathcal{L}_{\mathrm{IRM}}
        + \mathcal{L}_{\mathrm{FPR}}
        \right)$

        \State $\theta \leftarrow \theta - \eta \nabla_\theta \mathcal{L}$
    \EndFor
\EndFor
\end{algorithmic}
\end{algorithm}

\mypar{Design notes}
Although $K{=}G{=}6$, the partition head learns the mapping that
\emph{maximizes} $\mathcal{L}_{\text{REx}}$, not the identity map ---
it may merge groups that share the same spurious cue (e.g., Female-Dark
and Male-Dark) into one partition, concentrating the invariance penalty
where disparity is largest.
One BCE pre-training epoch (plus DP/FPR) is required before partition
learning starts; without it, the partition head maximizes a near-random
signal and collapses.
$\mathcal{L}_{\text{REx}}$ is excluded from Step~2 because its role is
partition discovery only; adding it to the feature update would produce
gradient signals that conflict with $\mathcal{L}_{\text{IRM}}$.
$\mathcal{L}_{\text{FPR}}$ operates on real images only ($y{=}0$), directly
targeting false-positive disparity, while $\mathcal{L}_{\text{DP}}$ acts on
all images for complementary prediction-rate coverage.

\subsection{Implementation Details}

\mypar{Architecture}
Xception~\cite{chollet2017xception} with ImageNet-pretrained weights and a
sigmoid output head.
Input images are resized to $299{\times}299$ and normalized to $[-1,1]$.
Data augmentation: random horizontal flip and color jitter (brightness,
contrast, saturation $\pm 0.2$).

\mypar{Optimizer}
Adam ($\text{lr}{=}10^{-5}$, weight decay $10^{-4}$) with CosineAnnealingLR
($T_{\max}{=}15$, $\eta_{\min}{=}0$).
Gradient norms clipped to 1.0.
BCE targets use label smoothing ($\epsilon{=}0.1$) to improve calibration.

\mypar{Hyperparameters}
$\lambda_{\text{IRM}}$ and $\lambda_{\text{REx}}$ are selected via a
grid search described in Section~\ref{sec:ablation}.
The values $\lambda_{\text{IRM}}{=}0.07$,
$\lambda_{\text{REx}}{=}5.0$ are selected as the configuration achieving the
optimal fairness score $S$ (Eq.~\ref{eq:fairness_score}) at deployed scale
($K{=}6$, 20k samples/8 epochs; see Section~\ref{sec:ablation} and
Supplementary Section~\ref{sec:supp_sweep} for the full grid).
Full training: $K{=}6$, $T{=}1$ partition step, 300\,000 samples/epoch,
batch size~32, 15~epochs; $\lambda{=}1.0$ throughout.

\section{Experiments}
\label{sec:experiments}

We evaluate IIL against ten baselines spanning standard, generalization-focused,
and fairness-aware detectors on the AI-Face benchmark.
All IIL hyperparameters except $\lambda_{\text{IRM}}$ and $\lambda_{\text{REx}}$
are fixed before training; these two are selected via the grid search in
Section~\ref{sec:ablation}.

\subsection{Dataset and Baselines}

\mypar{AI-Face}
The AI-Face dataset~\cite{aiface} contains over 1.3M face images labeled
Real or Fake with $G{=}6$ gender${\times}$skin-tone intersection labels
and predicted age labels (Child/Young/Adult/Middle/Senior).
We use the standard train/test split (${\approx}1.3$M training images,
329{,}309 test images, 10\,291 batches of size 32). Beyond this aggregate evaluation, we
also assess generalization on the four distinct constituent sources AI-Face
draws from --- Celeb-DF, DFD, DFDC, and FF++ (Section~\ref{sec:generalization},
Table~\ref{tab:crossdomain}) --- individually, not only in aggregate.

\mypar{Baselines}
We compare against all detectors evaluated in~\cite{aiface}:
standard detectors (Xception~\cite{chollet2017xception}, EfficientNet-B4~\cite{tan2019efficientnet}),
generalization detectors (F3Net~\cite{li2021f3net}, SPSL~\cite{liu2021spsl}, SRM~\cite{luo2021srm}, UCF~\cite{yan2023ucf},
CORE~\cite{ni2022core}),
and fairness-aware detectors (DAW-FDD~\cite{ju2024improving}, DAG-FDD~\cite{ju2024improving}, PG-FDD~\cite{lin2024pgfdd}).
Checkpoints are provided by~\cite{aiface}. Table~\ref{tab:results} reports
the baseline values published in Table~4 of~\cite{aiface} directly, for
exact comparability with prior work; we additionally re-evaluated all ten
released checkpoints ourselves under the same protocol as a consistency
check and found them to match the published values within minor
rounding-level differences (full 210-comparison breakdown in Supplementary Section~\ref{sec:supp_reeval}).

\subsection{Quantitative Results}

\begin{table*}[t]
\centering
\caption{Fairness and utility results on the AI-Face test set.
Benchmark values from Table~4 of~\cite{aiface}.
All values in~\%. $\downarrow$ lower is better; $\uparrow$ higher is better.
\textbf{Bold} = best across all 11 detectors.
EffB4\,=\,EfficientNet-B4; DAW\,=\,DAW-FDD; DAG\,=\,DAG-FDD.
\colorbox{iilgray}{Gray column} = IIL (ours).}
\label{tab:results}
\setlength{\tabcolsep}{3pt}
\resizebox{\linewidth}{!}{%
\small
\begin{tabular}{@{}ll rrrrrrrrrr>{\columncolor{iilgray}}r@{}}
\toprule
& & \multicolumn{11}{c}{\textbf{Detector}} \\
\cmidrule(lr){3-13}
\textbf{Attr.} & \textbf{Metric}
  & Xception~\cite{chollet2017xception} & EffB4~\cite{tan2019efficientnet}
  & F3Net~\cite{li2021f3net} & SPSL~\cite{liu2021spsl} & SRM~\cite{luo2021srm}
  & UCF~\cite{yan2023ucf} & CORE~\cite{ni2022core}
  & DAW~\cite{ju2024improving} & DAG~\cite{ju2024improving} & PG-FDD~\cite{lin2024pgfdd} & IIL (ours) \\
\midrule
\multirow{4}{*}{Skin Tone}
  & FMEO$\downarrow$ &  8.836 &  8.300 & 19.938 &  8.055 & 10.002 & 17.325 & 10.779 & 14.118 &  6.551 &  6.465 & \textbf{0.730} \\
  & FDP$\downarrow$  &  9.751 & \textbf{6.184} & 12.876 &  9.379 & 10.897 & 12.581 & 10.317 & 10.706 &  8.617 &  9.746 &  8.468 \\
  & FOAE$\downarrow$ &  1.271 &  4.377 &  2.818 &  1.135 &  0.915 &  1.883 &  1.332 &  1.667 &  1.388 &  0.882 & \textbf{0.770} \\
  & FEO$\downarrow$  & 12.132 & 11.062 & 23.708 &  9.789 & 14.239 & 21.920 & 13.069 & 16.604 &  7.383 &  9.115 & \textbf{1.537} \\
\cmidrule{1-13}
\multirow{4}{*}{Gender}
  & FMEO$\downarrow$ &  3.975 &  5.385 &  4.717 &  4.411 &  6.271 &  5.074 &  5.795 &  5.510 &  5.910 &  3.190 & \textbf{0.566} \\
  & FDP$\downarrow$  &  1.691 &  1.725 &  1.864 &  1.827 &  1.957 &  1.736 &  2.154 &  2.015 &  2.151 &  1.252 & \textbf{0.884} \\
  & FOAE$\downarrow$ &  0.975 &  1.487 &  1.129 &  1.037 &  1.772 &  1.451 &  1.389 &  1.325 &  1.420 &  1.071 & \textbf{0.329} \\
  & FEO$\downarrow$  &  4.143 &  5.863 &  4.870 &  4.534 &  6.780 &  5.510 &  5.931 &  5.696 &  6.066 &  3.702 & \textbf{0.836} \\
\cmidrule{1-13}
\multirow{4}{*}{Age}
  & FMEO$\downarrow$ & 27.883 &  6.796 & 38.801 & 27.614 & 24.843 & 47.500 & 33.882 & 45.466 & 15.229 & 14.804 & \textbf{1.134} \\
  & FDP$\downarrow$  & 10.905 & 11.849 & 14.906 & 11.232 & 11.570 & 17.049 & 12.564 & 14.106 & \textbf{9.633} & 10.467 & 14.345 \\
  & FOAE$\downarrow$ &  7.265 &  2.856 & 10.116 &  7.270 &  6.524 & 11.652 &  8.760 & 11.878 &  5.533 &  5.009 & \textbf{1.131} \\
  & FEO$\downarrow$  & 42.216 & 10.300 & 55.032 & 40.943 & 38.528 & 67.545 & 48.729 & 64.384 & 30.182 & 29.585 & \textbf{2.735} \\
\cmidrule{1-13}
\multirow{4}{*}{Intersection}
  & FMEO$\downarrow$ & 10.505 & 17.586 & 21.369 & 10.379 & 15.142 & 20.134 & 15.340 & 16.565 & 12.178 &  9.578 & \textbf{1.463} \\
  & FDP$\downarrow$  & 14.511 & \textbf{8.607} & 17.175 & 13.259 & 15.186 & 17.030 & 14.301 & 14.088 & 11.705 & 14.697 & 13.789 \\
  & FOAE$\downarrow$ &  2.536 &  8.461 &  4.870 &  2.464 &  3.998 &  3.536 &  2.775 &  3.547 &  4.035 &  3.062 & \textbf{1.553} \\
  & FEO$\downarrow$  & 24.315 & 25.114 & 47.783 & 21.679 & 30.112 & 43.376 & 28.840 & 33.122 & 26.295 & 18.348 & \textbf{3.867} \\
\midrule
\multirow{5}{*}{Utility}
  & AUC$\uparrow$  & 98.583 & 98.611 & 98.714 & 98.747 & 97.936 & 98.082 & 98.579 & 97.811 & 98.771 & 99.172 & \textbf{99.420} \\
  & ACC$\uparrow$  & 96.308 & 94.203 & 95.719 & 96.346 & 95.092 & 95.151 & 96.224 & 95.426 & 95.722 & 96.174 & \textbf{98.630} \\
  & AP$\uparrow$   & 99.350 & 99.542 & 99.453 & 99.356 & 99.172 & 99.273 & 99.360 & 99.015 & 99.498 & \textbf{99.694} &  99.610 \\
  & EER$\downarrow$&  5.149 &  6.689 &  5.256 &  4.371 &  6.483 &  7.708 &  5.145 &  7.063 &  5.499 &  4.961 & \textbf{1.570} \\
  & FPR$\downarrow$& 12.961 & 20.066 & 14.679 & 13.661 & 15.746 & 13.646 & 13.410 & 16.670 & 14.844 & 10.971 & \textbf{2.390} \\
\midrule
Metadata & Train epoch time &
    \textbf{1h15m} & 2h25m &
    1h18m & 1h20m & 3h10m &
    5h05m & 1h16m &
    1h25m & 1h17m & 7h20m & 1h30m \\
\bottomrule
\end{tabular}}
\end{table*}

Table~\ref{tab:results} shows that IIL achieves the best value on 17 of 21 metrics.
The four metrics where IIL does not lead are three FDP variants --- Skin FDP
(EffB4: 6.18\%), Inter FDP (EffB4: 8.60\%), Age FDP (DAG: 9.63\%) --- and AP.
FDP measures the prediction-rate gap between groups regardless of label;
EffB4's low Skin FDP reflects unusually balanced positive-prediction rates
rather than low FPR disparity, as evidenced by its skin FMEO of 8.30\%.
IIL's FPR parity loss and group-balanced sampling equalize per-group FPR
across the six gender${\times}$skin-tone partitions; this can shift overall
prediction rates in ways that widen demographic parity gaps for age groups,
which span the six partitions differently, explaining the higher Age FDP.
This is a structural property of the partition choice rather than a
post-hoc explanation: within-partition age-bracket composition varies
several-fold across the six training partitions (e.g.\ Child ranges
1.3\%--9.2\%, Middle-aged 10.7\%--35.5\%; full breakdown in
Supplementary Section~\ref{sec:supp_age_composition}), so equalizing FPR/DP across partitions does not
constrain FPR/DP within an age bracket that is unevenly distributed
across them.
On the utility side, PG-FDD's higher AP (99.69\% vs.\ 99.61\%) reflects a
slight ranking advantage at intermediate thresholds that is invisible in the
EER and FPR metrics.
The magnitude of IIL's FMEO gains is substantial: against PG-FDD, the
next-best baseline on three of four axes, intersection FMEO falls from
9.57\% to 1.46\% ($6.5{\times}$ reduction), gender FMEO from 3.19\% to
0.57\% ($5.6{\times}$), and skin FMEO from 6.465\% (DAG-FDD's 6.551\% a
close second) to 0.730\% ($8.9{\times}$); on age FMEO the next-best
baseline is EffB4, not PG-FDD, so the comparison is 6.796\% to 1.13\%
($6.0{\times}$).

\begin{figure*}[t]
\centering
\includegraphics[width=\linewidth]{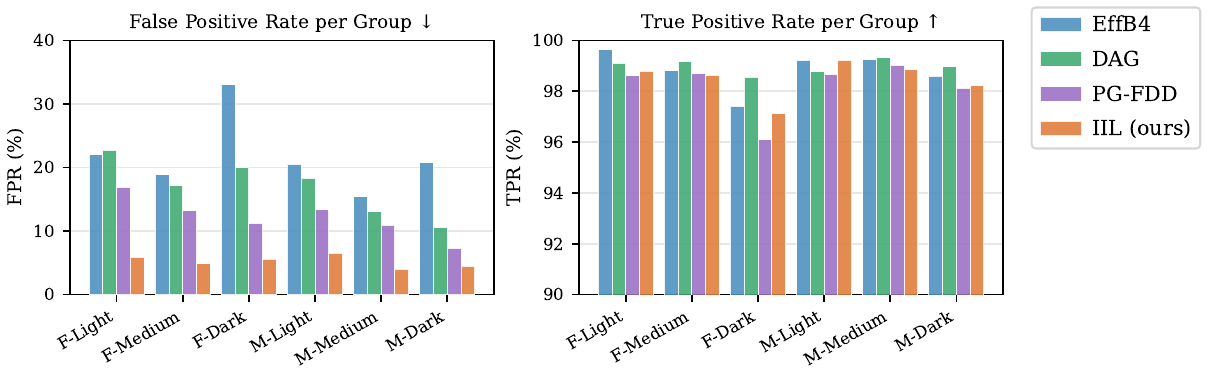}
\caption{Per-group FPR (left) and TPR (right) across the six gender${\times}$skin-tone
demographic partitions (Female/Male $\times$ Skin-Tone 1--3) for IIL (ours),
EfficientNet-B4, DAG-FDD, and PG-FDD.
IIL achieves near-uniform FPR across all six gender${\times}$skin-tone groups (max~3.4\%, range 1.5\,pp)
compared to EffB4 (max~33.0\%, range 24.1\,pp), DAG (max~22.8\%), and
PG-FDD (max~16.9\%), while maintaining TPR above 97\% for every group.
The dominant disparity axis in baselines is skin tone; gender contributes a
secondary gap within each skin-tone tier.}
\label{fig:per_group}
\end{figure*}

\begin{figure*}[t]
\centering
\begin{subfigure}[t]{0.48\linewidth}
  \centering
  \includegraphics[width=\linewidth,height=6cm,keepaspectratio]{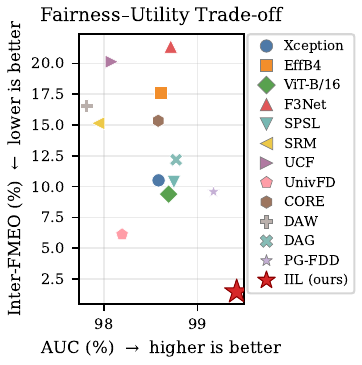}
  \caption{AUC vs.\ Intersection FMEO. IIL (red star) is Pareto-dominant:
  highest AUC (99.42\%) and lowest inter-FMEO (1.46\%) simultaneously;
  nearest competitor is $6.5{\times}$ worse on FMEO.}
  \label{fig:scatter}
\end{subfigure}%
\hfill
\begin{subfigure}[t]{0.48\linewidth}
  \centering
  \includegraphics[width=\linewidth,height=6cm,keepaspectratio]{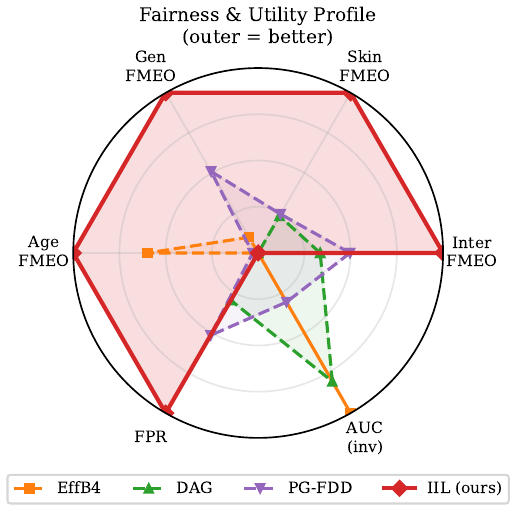}
  \caption{Normalized radar across AUC, FPR, inter-, skin-, gender-, and
  age-FMEO (outer rim = best). IIL fills uniformly; baselines trade one
  fairness axis against another.}
  \label{fig:radar}
\end{subfigure}
\caption{Fairness--utility analysis.}
\label{fig:scatter_radar}
\end{figure*}

\subsection{Hyperparameter Selection}
\label{sec:ablation}

$\lambda_{\text{IRM}}$ and $\lambda_{\text{REx}}$ are selected via a grid
search over 40 combinations ($8{\times}5$:
$\lambda_{\text{IRM}}\in[0.01,0.08]$, $\lambda_{\text{REx}}\in\{1\text{--}5\}$)
at $K{=}6$, 20k samples, 8 epochs --- close to the deployed regime
($K{=}6$, 300k samples, 15 epochs) while remaining tractable.
Configurations are ranked by:
\begin{equation}
\begin{split}
  S &= 2\cdot\mathrm{FMEO}_\mathrm{inter} + 2\cdot\mathrm{FMEO}_\mathrm{skin}\\
    &\quad + \mathrm{FMEO}_\mathrm{gen} + \mathrm{FMEO}_\mathrm{age} \quad(\downarrow).
\end{split}
\label{eq:fairness_score}
\end{equation}
Intersection and skin-tone FMEO receive double weight because they exhibit
the largest raw gaps in baseline detectors (up to 17.5\% and 8.3\%
respectively) and represent the primary axes of demographic disparity in
deepfake detection.
The selection $(\lambda_{\text{IRM}}{=}0.07,\,\lambda_{\text{REx}}{=}5.0)$
achieves the lowest $S{=}45.5$ and is used for all results below; full FPR
and fairness-score heatmaps over the $8{\times}5$ grid are given in
Supplementary Section~\ref{sec:supp_sweep}.

\subsection{Component Ablation}
\label{sec:component_ablation}

Table~\ref{tab:ablation_components} isolates the contribution of each IIL
component using the paper's hyperparameters
($\lambda{=}1.0$, $\lambda_{\mathrm{IRM}}{=}0.07$, $\lambda_{\mathrm{REx}}{=}5.0$, $K{=}6$)
on a 15-epoch run with 50\,000 training samples and 20\,000 held out for
evaluation --- the same epoch count as the full deployed run
(Table~\ref{tab:results}), at a reduced sample budget for tractability.
Variant~A is a plain BCE baseline; each subsequent variant adds one
component. Variant~E isolates the partition-learning step itself: $K{=}6$
with the group assignment fixed to the six annotated
gender$\times$skin-tone identities (Step~1 of Algorithm~1 skipped) rather
than learned via REx, otherwise identical to Variant~D.

\begin{table}[h]
\centering
\caption{Component ablation (15-epoch run, 50\,000 training samples,
  paper hyperparameters).
  $\checkmark$\,=\,enabled; $\times$\,=\,disabled; P$_{\mathrm{fix}}$\,=\,fixed
  identity partition instead of learned.
  $S$ = fairness score (Eq.~\ref{eq:fairness_score}; $\downarrow$).
  \textbf{Bold} = best non-collapsed value.
  $\dagger$collapsed: $\mathrm{FPR}{=}100\%$, all images predicted fake.}
\label{tab:ablation_components}
\resizebox{\columnwidth}{!}{%
\begin{tabular}{lccc cccc c}
\toprule
 & \multicolumn{3}{c}{\textbf{Components}} & & & & & \\
\cmidrule(lr){2-4}
\textbf{Variant} & \makecell{Balanced\\Sampler} & \makecell{FPR\\Loss} & \makecell{IRM/\\REx} &
  \textbf{AUC\,$\uparrow$} & \textbf{FPR\,$\downarrow$} &
  \textbf{inter-FMEO\,$\downarrow$} & \textbf{skin-FMEO\,$\downarrow$} & \textbf{$S$\,$\downarrow$} \\
\midrule
A & $\times$ & $\times$ & $\times$ & 98.23 & 12.18 & 17.18 & 6.58 & 54.12 \\
B & $\checkmark$ & $\times$ & $\times$ & 97.98 & 12.74 & 14.32 & 5.63 & 44.43 \\
C$^\dagger$ & $\checkmark$ & $\checkmark$ & $\times$ & 93.94 & 100.0 & --- & --- & --- \\
D & $\checkmark$ & $\checkmark$ & $\checkmark$ & \textbf{98.26} & \textbf{9.15} & \textbf{6.49} & \textbf{1.98} & \textbf{24.43} \\
E ($P_{\mathrm{fix}}$) & $\checkmark$ & $\checkmark$ & $\checkmark$ & 98.22 & 9.68 & 6.97 & 3.39 & 26.62 \\
\bottomrule
\end{tabular}}
\end{table}

At $\lambda{=}1.0$, adding only the FPR parity loss (Variant~C) causes
training collapse: the model predicts every image as fake ($\mathrm{FPR}{=}100\%$),
trivially equalizing group rates.
Full IIL (D) avoids collapse and achieves the best AUC (98.26\%), FPR
(9.15\%), inter-FMEO (6.49\%), skin-FMEO (1.98\%), and fairness score
($S{=}24.43$) simultaneously --- a Pareto win with no utility trade-off at
this scale, not merely a fairness gain purchased with accuracy. This
demonstrates that IRM/REx is a necessary stabilizer for the parity losses,
not merely an additive component: without it (C), the same parity losses
collapse training; with it (D), they combine into the best result on every
axis measured. Learned partitioning (D) also outperforms the fixed-identity
alternative (E) on every metric, showing the REx-learned worst-case
partitions are more informative than the raw annotated groups themselves.
In full training (300\,000 samples/epoch), this compounding effect
drives IIL to inter-FMEO of 1.46\%; see Table~\ref{tab:results}. Whether
Variant~C's collapse is an artifact of this reduced sample budget, and
learning curves bridging this ablation-scale run and the full-scale
deployment, are given in Supplementary Section~\ref{sec:supp_bridging}.

\begin{figure*}[t]
\centering
\includegraphics[width=\linewidth]{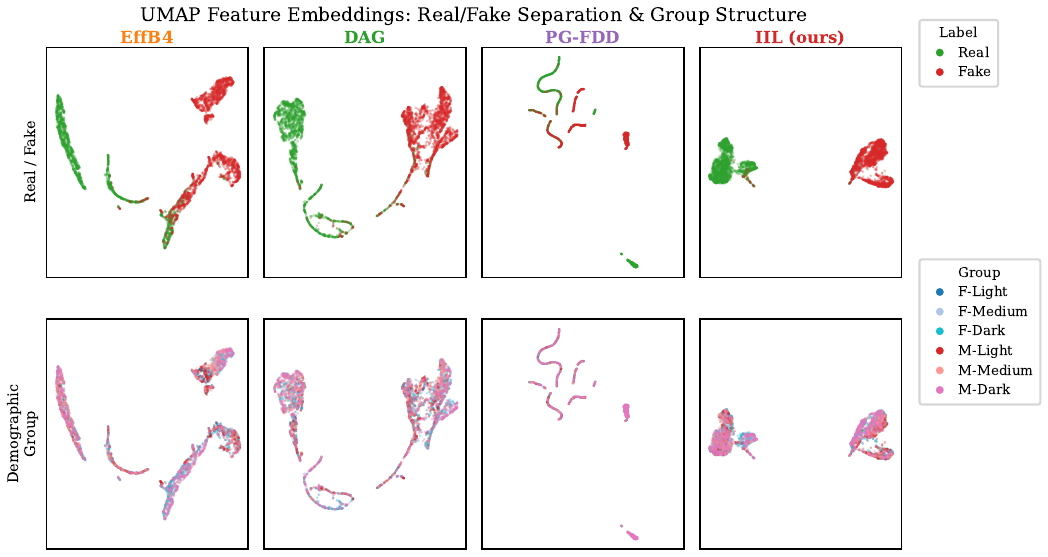}
\caption{UMAP~\cite{mcinnes2018umap} embeddings for EffB4, DAG, PG-FDD, and IIL (ours).
\textbf{Top row}: colored by Real/Fake label.
EffB4 spreads into multiple elongated thread-like strands; DAG produces two
large compact blobs; PG-FDD yields serpentine filamentous curves; IIL achieves
the tightest compact clusters with the sharpest Real/Fake boundary.
All four models achieve clear Real/Fake separation, consistent with AUC values above 98\%.
\textbf{Bottom row}: colored by intersectional demographic group.}
\label{fig:umap}
\end{figure*}

\begin{figure*}[t]
\centering
\includegraphics[width=\linewidth]{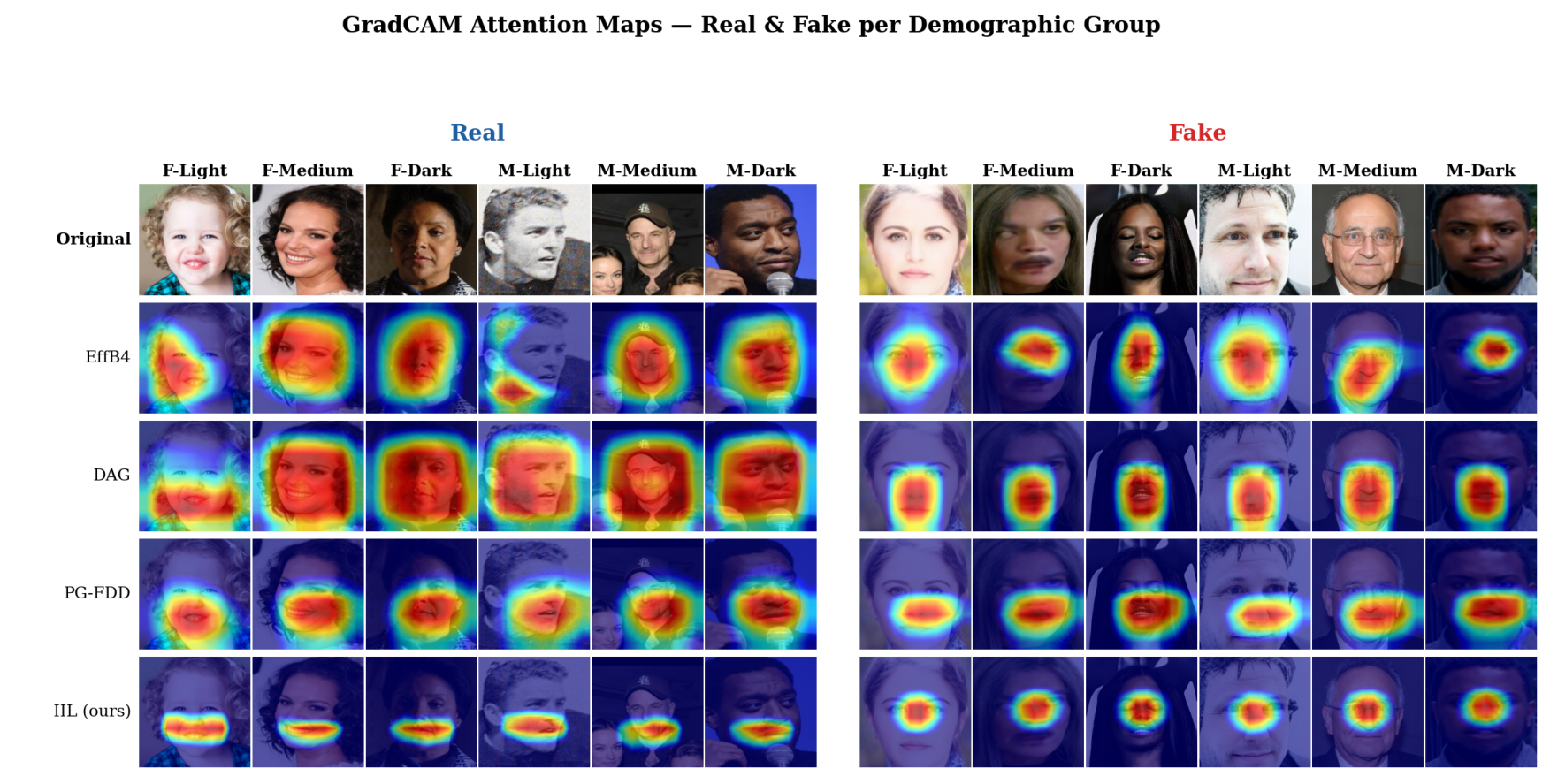}
\caption{GradCAM~\cite{selvaraju2017gradcam} attention maps across the six
gender${\times}$skin-tone groups and four models. Left block: real images;
right block: fake images.}
\label{fig:attention}
\end{figure*}

\subsection{Analysis}
\label{sec:analysis}

\mypar{Per-group rates (Figure~\ref{fig:per_group})}
Baselines show a clear skin-tone-driven disparity: Female-Dark and Male-Dark
attract the highest FPR (up to 33.0\% for EffB4, 22.8\% for DAG), with an
FPR range of 24.1\,pp for EffB4 versus just 1.5\,pp for IIL.
IIL caps all six groups below 3.4\% FPR while maintaining TPR above 97\%,
confirming minority-group accuracy is lifted rather than majority-group accuracy degraded.

\mypar{Fairness--utility trade-off (Figure~\ref{fig:scatter})}
Standard detectors cluster at AUC 97.9--98.7\% with inter-FMEO 10--21\%;
fairness-aware baselines (DAW, DAG, PG-FDD) span 9.5--16.6\%.
IIL is Pareto-dominant: highest AUC (99.42\%) and lowest inter-FMEO (1.46\%)
simultaneously, $6.5{\times}$ below the nearest competitor (PG-FDD, 9.58\%).
The radar (Figure~\ref{fig:radar}) confirms the advantage is uniform across
all six axes; fairness-aware baselines gain on one axis at the cost of another.

\mypar{Training dynamics (Supplementary Section~\ref{sec:supp_bridging})}
FPR drops from 11.2\% to below 5\% by epoch~3, driven by $\mathcal{L}_{\text{FPR}}$,
which penalizes false positives on real images before IRM and REx have reshaped
the representation.
Inter-FMEO oscillates at epochs~4--7 because the partition head is reassigning
groups as the encoder changes --- a transient effect that settles by epoch~9.
AUC stays above 99.3\% throughout, confirming fairness and detection do not trade off.

\mypar{UMAP feature geometry (Figure~\ref{fig:umap})}
The top row shows clear Real/Fake separation in all four models, consistent
with AUC values above 98\%.
In the demographic bottom row, pinkish and bluish hues occupy visibly separate
regions in EffB4 and DAG, indicating their representations encode demographic
identity and expose the classifier to group-correlated signals
(inter-FMEO 17.5\% and 12.1\%).
PG-FDD's demographic colors are the most compact and co-located of the
four, consistent with its representation carrying the least group
information. IIL's demographic colors are also more mixed than
EffB4/DAG in this 2D projection, but this visual compactness should not
be read as evidence that group identity is absent from IIL's features:
UMAP's nonlinear, neighbor-preserving projection can visually blend
structure that remains linearly recoverable in the full feature space.
A quantitative demographic-group classifiability probe (Supplementary Section~\ref{sec:supp_invariance_probe})
confirms this: IIL's features are only moderately harder to decode group
from than DAG-FDD's (41.1\% vs.\ 37.7\% 6-way group accuracy, a linear
probe on 50-PCA-component features; majority-class baseline 17.0\%), well
above PG-FDD's 22.1\%. IIL's fairness gains (Table~\ref{tab:results}) come
from equalizing the classifier's risk across partitions
(Section~\ref{sec:method}), not from removing demographic information from
the representation itself --- IRM/REx (Eqs.~\ref{eq:irm}--\ref{eq:rex})
constrain what the classifier does with a partition-specific cue, not
whether that cue remains present in $\Phi(x)$.

\mypar{Attention maps (Figure~\ref{fig:attention})}
On real images, EffB4 and DAG's activation spreads broadly across facial
skin regardless of demographic group --- skin texture acting as a
spurious ``fakeness'' cue rather than a landmark-anchored one.
IIL shows a clear real/fake asymmetry, but not a diffuse one: on
\emph{real} images its activation is a narrow horizontal band anchored to
the mouth/lip region; on \emph{fake} images it collapses to a smaller,
rounder blob centered on the nose--mouth junction, consistent with its
near-zero skin-FMEO (0.73\%).
This mouth-band/nose-blob asymmetry is comparable in size and position
across all six groups and reproduces across independently resampled
image sets, unlike the baseline pattern.
(Single sample per group in this figure; not averaged over multiple
images per cell.)

\mypar{Checkpoint selection}
Epoch~13 achieves the best $S{=}6.09$.
For epoch~14, we recorded a lower raw FPR (2.29\% vs.\ 2.39\%) and
slightly better utility overall, but a regression on all four gender and
age metrics and three of four intersection metrics ($S{=}6.35$), winning
only 9 of 21 metrics. Hence, epoch~13 was chosen as the best checkpoint.

\FloatBarrier
\subsection{Generalization Across Datasets}
\label{sec:generalization}

\begin{table*}[!t]
\centering
\caption{Generalization across datasets. Evaluated on AI-Face
  demographically annotated subsets~\cite{aiface} (8K--36K images per dataset).
  AUC$\uparrow$ and FPR$\downarrow$ (\%); \textbf{FMEO$\downarrow$} (\%) is the
  intersectional Max Equalized Odds gap across the six gender${\times}$skin-tone
  groups, as defined in \S\ref{sec:metrics}.
  \textbf{Bold} = best per metric.}
\label{tab:crossdomain}
\resizebox{\linewidth}{!}{%
\begin{tabular}{lrrrrrrrrrrrr}
\toprule
& \multicolumn{3}{c}{\textbf{Celeb-DF}} & \multicolumn{3}{c}{\textbf{DFD}}
& \multicolumn{3}{c}{\textbf{DFDC}} & \multicolumn{3}{c}{\textbf{FF++}} \\
\cmidrule(lr){2-4}\cmidrule(lr){5-7}\cmidrule(lr){8-10}\cmidrule(lr){11-13}
\textbf{Model}
  & AUC$\uparrow$ & FPR$\downarrow$ & FMEO$\downarrow$
  & AUC$\uparrow$ & FPR$\downarrow$ & FMEO$\downarrow$
  & AUC$\uparrow$ & FPR$\downarrow$ & FMEO$\downarrow$
  & AUC$\uparrow$ & FPR$\downarrow$ & FMEO$\downarrow$ \\
\midrule
Xception~\cite{chollet2017xception}
  & 92.85 & 28.75 & 87.50
  & 85.42 & 31.40 & 16.80
  & 75.24 & 13.56 & 61.35
  & 86.57 & 17.33 & 82.18 \\
EffB4~\cite{tan2019efficientnet}
  & 71.85 & 85.34 & 21.74
  & 72.98 & 87.11 & 16.69
  & 64.90 & 55.06 & 81.71
  & 71.55 & 35.04 & 75.79 \\
F3Net~\cite{li2021f3net}
  & 89.06 &  9.26 & 93.33
  & 81.82 & 16.30 & 23.75
  & 73.54 &  9.03 & 47.46
  & 81.84 &  6.03 & 49.89 \\
SPSL~\cite{liu2021spsl}
  & 93.46 & 14.47 & 87.50
  & 84.25 & 51.62 & 10.18
  & 76.66 & 15.16 & 68.12
  & 90.66 & 17.12 & 69.09 \\
SRM~\cite{luo2021srm}
  & 87.64 &  5.40 & 37.50
  & 84.60 & 19.76 & 12.91
  & 81.67 & 16.49 & 71.43
  & 81.75 & 26.22 & 59.49 \\
UCF~\cite{yan2023ucf}
  & 90.10 &  5.65 & 12.30
  & 88.65 & 16.47 &  9.14
  & 79.90 & 17.60 & 80.00
  & 90.70 &  5.51 & 58.25 \\
CORE~\cite{ni2022core}
  & 90.68 & 20.98 & 87.50
  & 80.32 & 37.99 & 34.22
  & 72.02 & 24.57 & 71.43
  & 84.72 & 20.57 & 78.08 \\
\midrule
DAW~\cite{ju2024improving}
  & 90.46 & 20.71 & 36.79
  & 86.59 & 37.31 & 41.03
  & 79.27 & 27.71 & 78.74
  & 88.48 &  8.55 & 65.79 \\
DAG~\cite{ju2024improving}
  & 89.39 & 13.38 & 91.46
  & 82.45 & 27.09 & 17.98
  & 72.19 & 45.45 & 69.97
  & 85.86 & 10.35 & 79.12 \\
PG-FDD~\cite{lin2024pgfdd}
  & 93.63 & \textbf{1.45} & 12.50
  & 84.29 & 8.75 & 28.17
  & 78.78 & 13.82 & 62.25
  & 87.20 & \textbf{2.65} & 56.34 \\
\midrule
\rowcolor{iilgray}
IIL (ours)
  & \textbf{98.86} &  6.57 &  \textbf{7.83}
  & \textbf{97.00} & \textbf{8.57} & \textbf{5.55}
  & \textbf{95.60} & \textbf{7.03} & \textbf{14.67}
  & \textbf{96.50} & 13.10 & \textbf{24.11} \\
\bottomrule
\end{tabular}}
\end{table*}

Table~\ref{tab:crossdomain} evaluates all models on four datasets:
Celeb-DF~\cite{li2020celebdf}, DFD~\cite{google2019dfd},
DFDC~\cite{dolhansky2020dfdc}, and FF++~\cite{rossler2019faceforensics}.
As AI-Face aggregates images from these same sources, IIL's results here
reflect held-out partition generalization --- disjoint splits of the same
underlying distribution --- rather than true domain transfer.
All metrics are computed on the demographically annotated subsets
provided by~\cite{aiface} (8K--36K images per dataset).

Table~\ref{tab:crossdomain} shows that IIL achieves the highest AUC on all
four datasets (98.86\%, 97.00\%, 95.60\%, 96.50\%), outperforming all baselines
by a substantial margin --- roughly 5\,pp above the next-best on Celeb-DF and
nearly 14\,pp on DFDC.
IIL attains the lowest FMEO on all four datasets: 7.83\% on Celeb-DF
(UCF: 12.30\%), 5.55\% on DFD (UCF: 9.14\%), 14.67\% on DFDC (F3Net:
47.46\%, $3.2{\times}$ lower), and 24.11\% on FF++ (PG-FDD: 56.34\%,
$2.3{\times}$ lower). On raw FPR, IIL also leads on DFD (8.57\%) and
DFDC (7.03\%); the exception is Celeb-DF and FF++, where PG-FDD's
lower FPR (1.45\%, 2.65\%) comes with a far wider FMEO gap (12.50\%,
56.34\%) than IIL's (7.83\%, 24.11\%).

\section{Conclusion}
\label{sec:conclusion}

This research indicates that demographic failures in deepfake detection
models stem from the models learning skin texture as a substitute for
identifying forgeries.
This phenomenon is a representation-level pathology that cannot be solved
by output-level demographic parity penalties.
Intersectional Invariant Learning (IIL) strives to address this issue.
By enforcing IRMv1~\cite{arjovsky2019irm} across the most unfavorable
demographic divisions, IIL seeks to provide a representation that is
skin-texture cue free.
Using GradCAM, it is shown that IIL's attention is real/fake asymmetric:
a narrow band on the mouth/lip region for real images versus a compact
blob on the nose--mouth junction for fake images, a pattern that
reproduces across independently resampled images and stays comparable in
size and position across all six demographic groups, unlike baselines'
broad, skin-spanning activation on real images.

On the AI-Face benchmark~\cite{aiface}, IIL performs Pareto-dominant,
achieving the highest AUC (99.42\%) and the lowest intersectional FMEO
(1.46\%), with $4.6{\times}$ and $6.5{\times}$ improvements over the
previous state-of-the-art fairness-aware detector
(PG-FDD~\cite{lin2024pgfdd}: FPR 10.97\%, inter-FMEO 9.58\%).
Using UMAP, the feature spaces of the six gender${\times}$skin-tone groups
were shown to be visually more mixed than EffB4/DAG's, though not more
than PG-FDD's, and the metric improvements themselves come from
equalizing the classifier's risk across partitions rather than from
removing demographic information from the representation
(Section~\ref{sec:analysis}).
IIL achieves the best performance across all four protected attributes
(Skin Tone, Gender, Age, and Intersection), leading on 17 of 21 metrics,
with the remaining four being three FDP variants~\cite{dwork2012fairness}
(Skin, Age, and Intersection FDP) and Average Precision (AP).

\mypar{Broader significance}
Adding DP penalties~\cite{agarwal2018reductions} to existing detectors is
insufficient: models that equalize prediction rates may still impose
asymmetric false-positive burdens across demographic groups~\cite{hardt2016equality}.
Invariant learning~\cite{arjovsky2019irm,krueger2021rex} constrains what
the representation encodes, aligning the inductive bias with the fairness
criteria that matter.

\mypar{Future directions}
To improve age-attribute fairness evaluation, we consider exploring, in future, the use of verified ground-truth age labels in place of predicted ones~\cite{aiface}.
To achieve more granular invariance penalties, we will explore soft or hierarchical partition assignments instead of hard $K{=}6$ clusters.
To close the demographic-decodability gap flagged in Supplementary Section~\ref{sec:supp_invariance_probe} without sacrificing IIL's risk-level fairness gains, we plan to explore an explicit feature-level invariance penalty (e.g.\ distribution matching or adversarial debiasing) alongside IRM/REx.
All results in this paper use a single backbone (Xception~\cite{chollet2017xception}); we plan to evaluate IIL with additional architectures to observe performance across backbones.

{\small
\bibliographystyle{ieeenat_fullname}
\bibliography{main}
}

\clearpage
\appendix
\twocolumn[{\centering{\Large\bf Supplementary Material\par}\vspace{1.5em}}]
\section{Full-Test-Split Baseline Re-evaluation}
\label{sec:supp_reeval}

\S5.1/Table~2 reports baseline values published in Table~4
of~\cite{aiface} directly rather than from our own runs, for exact
comparability with prior work. This section provides the independent
re-evaluation of the released checkpoints, confirming agreement with
those published numbers.
Table~\ref{tab:supp_reeval} reports, for all ten baselines plus IIL,
every metric in Table~2 (FMEO/FDP/FOAE/FEO across Skin Tone, Gender,
Age, and Intersection, plus AUC/ACC/AP/EER/FPR) computed by us on the
full AI-Face test split ($n{=}329{,}309$). Layout matches Table~2
exactly (same row/column order) so the two can be compared directly;
aggregate deviation from the published values is summarized in the
caption rather than annotated per cell. Full protocol in
\S\ref{sec:supp_reeval_protocol}.

\begin{table*}[t]
\centering
\caption{Full-test-split re-evaluation of all ten baselines and IIL,
same layout as Table~2. \textbf{Bold} = best per row. Deviation from
Table~2's published baseline values (210 cells: 10 baselines
$\times$ 21 metrics) is rounding-level throughout: max $|\Delta|$ =
0.772\,pp (Intersection FEO, SPSL), mean $|\Delta|$ = 0.026\,pp
overall. Per-attribute max/mean $|\Delta|$: Skin Tone 0.298/0.021,
Gender 0.077/0.009, Age 0.615/0.053, Intersection 0.772/0.036 (N=40
each), Utility 0.027/0.003 (N=50). No cell deviates by more than
0.8\,pp, confirming rounding-level agreement with the published
values.}
\label{tab:supp_reeval}
\resizebox{\linewidth}{!}{%
\begin{tabular}{@{}ll rrrrrrrrrrr@{}}
\toprule
& & \multicolumn{11}{c}{\textbf{Detector}} \\
\cmidrule(lr){3-13}
\textbf{Attr.} & \textbf{Metric} & Xception & EffB4 & F3Net & SPSL & SRM & UCF & CORE & DAW & DAG & PG-FDD & IIL \\
\midrule
  \multirow{4}{*}{Skin Tone} & FMEO$\downarrow$ & 8.851 & 8.276 & 19.935 & 8.353 & 9.998 & 17.321 & 10.775 & 14.122 & 6.555 & 6.468 & \textbf{0.730} \\
   & FDP$\downarrow$ & 9.759 & \textbf{6.195} & 12.876 & 9.479 & 10.894 & 12.582 & 10.318 & 10.707 & 8.618 & 9.747 & 8.468 \\
   & FOAE$\downarrow$ & 1.270 & 4.367 & 2.814 & 1.163 & 0.911 & 1.878 & 1.329 & 1.669 & 1.388 & 0.881 & \textbf{0.770} \\
   & FEO$\downarrow$ & 12.144 & 11.061 & 23.703 & 10.049 & 14.233 & 21.921 & 13.068 & 16.607 & 7.389 & 9.120 & \textbf{1.537} \\
\cmidrule{1-13}
  \multirow{4}{*}{Gender} & FMEO$\downarrow$ & 3.978 & 5.366 & 4.707 & 4.488 & 6.264 & 5.076 & 5.805 & 5.507 & 5.912 & 3.188 & \textbf{0.566} \\
   & FDP$\downarrow$ & 1.694 & 1.721 & 1.863 & 1.848 & 1.955 & 1.737 & 2.155 & 2.014 & 2.150 & 1.250 & \textbf{0.884} \\
   & FOAE$\downarrow$ & 0.973 & 1.482 & 1.125 & 1.053 & 1.770 & 1.451 & 1.393 & 1.324 & 1.422 & 1.071 & \textbf{0.329} \\
   & FEO$\downarrow$ & 4.144 & 5.843 & 4.858 & 4.608 & 6.774 & 5.511 & 5.943 & 5.693 & 6.070 & 3.701 & \textbf{0.836} \\
\cmidrule{1-13}
  \multirow{4}{*}{Age} & FMEO$\downarrow$ & 27.823 & 6.759 & 38.804 & 28.169 & 24.833 & 47.564 & 33.913 & 45.469 & 15.231 & 14.806 & \textbf{1.134} \\
   & FDP$\downarrow$ & 10.895 & 11.869 & 14.897 & 11.375 & 11.570 & 17.076 & 12.581 & 14.106 & \textbf{9.633} & 10.467 & 14.345 \\
   & FOAE$\downarrow$ & 7.243 & 2.857 & 10.124 & 7.414 & 6.524 & 11.670 & 8.766 & 11.878 & 5.533 & 5.026 & \textbf{1.131} \\
   & FEO$\downarrow$ & 42.136 & 10.246 & 55.027 & 41.558 & 38.508 & 67.644 & 48.729 & 64.376 & 30.188 & 29.621 & \textbf{2.735} \\
\cmidrule{1-13}
  \multirow{4}{*}{Intersection} & FMEO$\downarrow$ & 10.522 & 17.554 & 21.357 & 10.676 & 15.137 & 20.122 & 15.340 & 16.565 & 12.184 & 9.584 & \textbf{1.463} \\
   & FDP$\downarrow$ & 14.511 & \textbf{8.619} & 17.175 & 13.350 & 15.183 & 17.033 & 14.301 & 14.091 & 11.705 & 14.700 & 13.789 \\
   & FOAE$\downarrow$ & 2.536 & 8.451 & 4.863 & 2.477 & 3.992 & 3.529 & 2.780 & 3.547 & 4.037 & 3.064 & \textbf{1.553} \\
   & FEO$\downarrow$ & 24.331 & 25.119 & 47.775 & 22.451 & 30.077 & 43.382 & 28.863 & 33.131 & 26.307 & 18.351 & \textbf{3.867} \\
\cmidrule{1-13}
\multirow{5}{*}{Utility}
  & AUC$\uparrow$ & 98.585 & 98.612 & 98.713 & 98.749 & 97.935 & 98.080 & 98.579 & 97.809 & 98.772 & 99.172 & \textbf{99.420} \\
  & ACC$\uparrow$ & 96.308 & 94.211 & 95.715 & 96.345 & 95.091 & 95.146 & 96.221 & 95.427 & 95.723 & 96.174 & \textbf{98.630} \\
  & AP$\uparrow$ & 99.352 & 99.543 & 99.453 & 99.358 & 99.171 & 99.272 & 99.361 & 99.014 & 99.498 & \textbf{99.694} & 99.610 \\
  & EER$\downarrow$ & 5.145 & 6.686 & 5.258 & 4.375 & 6.483 & 7.710 & 5.142 & 7.063 & 5.498 & 4.961 & \textbf{1.570} \\
  & FPR$\downarrow$ & 12.949 & 20.039 & 14.679 & 13.658 & 15.752 & 13.654 & 13.413 & 16.666 & 14.843 & 10.970 & \textbf{2.390} \\
\bottomrule
\end{tabular}
}
\end{table*}

\subsection{Re-evaluation Protocol}
\label{sec:supp_reeval_protocol}

We evaluate all ten released baseline checkpoints plus IIL on the full
AI-Face test split (329{,}309 images spanning all four source
categories: Real, GANs, DMs, and deepfakes), using each detector's own
training-time preprocessing (each baseline is resized/normalized per
its own architecture's convention, $256{\times}256$ with
$\mathrm{mean}{=}\mathrm{std}{=}0.5$ across all ten, matching the
AI-Face benchmark's own evaluation code exactly; the resulting
re-evaluated values are given in Table~\ref{tab:supp_reeval}).
Demographic groupings follow
the benchmark's own convention: the \texttt{Intersection} column maps
to the six gender$\times$skin-tone cells, \texttt{Predicted Age} to
five age buckets (Child/Young/Adult/Middle/Senior). FMEO/FDP/FOAE/FEO
are computed via the benchmark's own \texttt{acc\_fairness}
implementation (not a re-derivation) for exact comparability with the
published Table~4 of~\cite{aiface}.

\section{Hyperparameter Sweep at Deployed Scale}
\label{sec:supp_sweep}

$\lambda_{\text{IRM}}$ and $\lambda_{\text{REx}}$ (\S5.3) are
selected via a grid search over 40 combinations
($8{\times}5$: $\lambda_{\text{IRM}} \in [0.01, 0.08]$,
$\lambda_{\text{REx}} \in \{1\text{--}5\}$) at $K{=}6$ (matching deployed),
20k samples, 8 epochs --- close to the deployed regime ($K{=}6$, 300k
samples, 15 epochs) while remaining tractable (one full-scale training
run alone takes ${\sim}22$h, so a 40-combination grid at deployed scale
would take ${\sim}37$ days).

\begin{figure*}[t]
\centering
\includegraphics[width=0.4\linewidth]{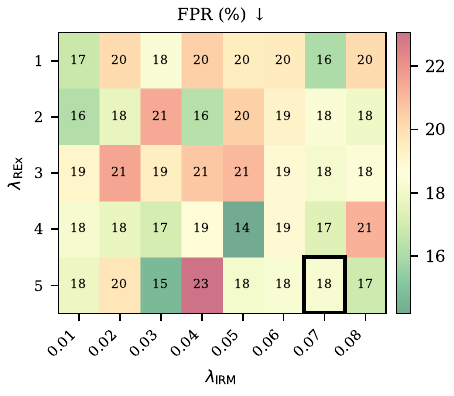}
\includegraphics[width=0.4\linewidth]{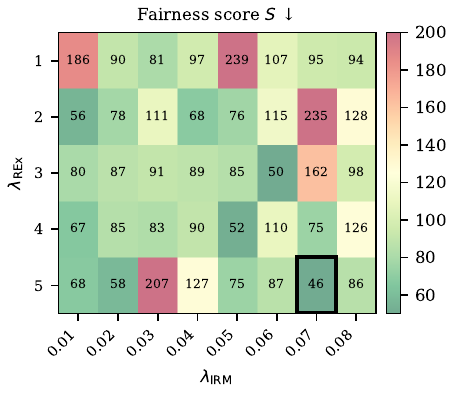}
\caption{FPR (left) and fairness score $S$ (right) across the full
$8\times5$ grid at $K{=}6$/20k samples/8 epochs. Black box = selected
point $(\lambda_{\text{IRM}}{=}0.07,\lambda_{\text{REx}}{=}5.0)$.}
\label{fig:supp_sweep_heatmap}
\end{figure*}

Figure~\ref{fig:supp_sweep_heatmap} shows the resulting grid (the main
paper's \S4 states the selected point and score numerically; the figure
itself lives only here to avoid duplicating it in both documents).
Table~\ref{tab:supp_sweep} gives the full per-combination breakdown, all
40 rows sorted by fairness score $S$ (best first).

The selected point, $(\lambda_{\text{IRM}}{=}0.07,\lambda_{\text{REx}}{=}5.0)$,
achieves the lowest $S{=}45.5$.

\begin{table}[h]
\centering
\caption{Full $8\times5$ sweep grid at $K{=}6$/20k samples/8 epochs, all
40 combinations, sorted by fairness score $S$ (Eq.~3;
\textbf{bold} = selected point, lowest $S$). AUC/ACC/FPR in \%; FMEO
columns in \% (inter/skin/gen/age, left to right).}
\label{tab:supp_sweep}
\resizebox{\linewidth}{!}{%
\begin{tabular}{lrrrrrrrr}
\toprule
\textbf{$(\lambda_{\text{IRM}},\lambda_{\text{REx}})$} & \textbf{$S\downarrow$} & \textbf{AUC$\uparrow$} & \textbf{ACC$\uparrow$} & \textbf{FPR$\downarrow$} & \textbf{inter} & \textbf{skin} & \textbf{gen} & \textbf{age} \\
\midrule
(0.07, 5.00) & \textbf{45.5} & 95.49 & 90.58 & 18.28 & 10.89 & 5.09 & 4.96 & 8.62 \\
(0.06, 3.00) & 50.3 & 96.24 & 90.94 & 18.90 & 11.71 & 5.70 & 3.58 & 11.86 \\
(0.05, 4.00) & 52.3 & 96.32 & 91.10 & 14.18 & 13.14 & 6.52 & 1.11 & 11.87 \\
(0.01, 2.00) & 55.6 & 95.93 & 90.48 & 16.17 & 13.65 & 7.92 & 2.66 & 9.82 \\
(0.02, 5.00) & 58.1 & 95.44 & 90.66 & 19.75 & 14.27 & 7.10 & 3.79 & 11.55 \\
(0.01, 4.00) & 66.6 & 96.22 & 91.40 & 18.23 & 23.08 & 4.48 & 2.58 & 8.89 \\
(0.01, 5.00) & 67.7 & 96.66 & 91.80 & 17.90 & 21.66 & 1.64 & 7.12 & 14.02 \\
(0.04, 2.00) & 68.2 & 95.85 & 90.80 & 16.27 & 20.00 & 8.55 & 3.06 & 8.01 \\
(0.05, 5.00) & 74.7 & 96.11 & 91.36 & 18.16 & 19.75 & 6.46 & 8.31 & 14.02 \\
(0.07, 4.00) & 75.1 & 95.40 & 90.74 & 17.42 & 25.00 & 6.04 & 1.45 & 11.60 \\
(0.05, 2.00) & 76.5 & 96.18 & 91.20 & 20.42 & 25.00 & 2.72 & 5.16 & 15.87 \\
(0.02, 2.00) & 78.0 & 96.43 & 91.20 & 17.69 & 19.30 & 11.97 & 5.50 & 10.00 \\
(0.01, 3.00) & 80.1 & 95.98 & 91.14 & 19.12 & 21.82 & 10.71 & 3.73 & 11.31 \\
(0.03, 1.00) & 80.7 & 96.48 & 91.68 & 18.38 & 24.91 & 10.72 & 0.51 & 8.90 \\
(0.03, 4.00) & 82.6 & 96.13 & 90.28 & 17.09 & 22.00 & 6.79 & 7.86 & 17.19 \\
(0.05, 3.00) & 84.6 & 95.79 & 91.28 & 21.03 & 26.54 & 8.87 & 5.37 & 8.37 \\
(0.02, 4.00) & 84.6 & 96.21 & 91.12 & 17.76 & 20.00 & 13.25 & 6.83 & 11.25 \\
(0.08, 5.00) & 86.1 & 96.16 & 90.74 & 16.90 & 26.24 & 6.68 & 8.25 & 12.06 \\
(0.06, 5.00) & 86.5 & 96.15 & 90.90 & 18.31 & 25.00 & 8.32 & 4.73 & 15.16 \\
(0.02, 3.00) & 87.3 & 95.67 & 91.06 & 21.47 & 25.00 & 11.11 & 5.45 & 9.66 \\
(0.04, 3.00) & 88.6 & 95.78 & 90.54 & 20.64 & 28.57 & 8.33 & 3.65 & 11.11 \\
(0.04, 4.00) & 90.4 & 96.44 & 91.62 & 18.71 & 33.33 & 5.45 & 4.96 & 7.88 \\
(0.02, 1.00) & 90.4 & 96.05 & 91.18 & 20.07 & 23.20 & 14.84 & 3.24 & 11.10 \\
(0.03, 3.00) & 90.8 & 96.44 & 91.62 & 19.48 & 22.99 & 11.68 & 2.88 & 18.60 \\
(0.08, 1.00) & 94.1 & 95.72 & 91.06 & 20.06 & 27.06 & 12.62 & 4.13 & 10.58 \\
(0.07, 1.00) & 95.3 & 96.31 & 90.96 & 16.06 & 21.68 & 17.56 & 4.97 & 11.81 \\
(0.04, 1.00) & 97.0 & 95.99 & 91.06 & 20.44 & 33.33 & 5.51 & 5.36 & 13.94 \\
(0.08, 3.00) & 98.0 & 96.00 & 90.82 & 18.50 & 21.33 & 19.44 & 2.62 & 13.82 \\
(0.06, 1.00) & 107.2 & 96.16 & 91.36 & 19.53 & 23.65 & 20.77 & 6.98 & 11.34 \\
(0.06, 4.00) & 110.5 & 95.67 & 91.02 & 18.89 & 23.73 & 19.69 & 7.97 & 15.64 \\
(0.03, 2.00) & 110.7 & 96.06 & 91.54 & 21.40 & 40.00 & 6.30 & 6.82 & 11.30 \\
(0.06, 2.00) & 114.7 & 95.96 & 90.94 & 19.24 & 40.00 & 10.08 & 4.65 & 9.91 \\
(0.08, 4.00) & 126.4 & 95.74 & 91.12 & 21.18 & 28.37 & 23.56 & 9.24 & 13.32 \\
(0.04, 5.00) & 127.2 & 95.53 & 90.96 & 23.06 & 43.33 & 13.50 & 5.04 & 8.49 \\
(0.08, 2.00) & 128.1 & 96.53 & 91.18 & 17.97 & 50.00 & 8.52 & 1.28 & 9.82 \\
(0.07, 3.00) & 162.4 & 95.84 & 90.84 & 18.19 & 43.05 & 32.96 & 2.41 & 8.02 \\
(0.01, 1.00) & 185.6 & 96.45 & 90.72 & 16.75 & 60.00 & 24.08 & 5.39 & 12.10 \\
(0.03, 5.00) & 206.5 & 95.77 & 89.76 & 14.69 & 60.00 & 35.79 & 3.90 & 11.06 \\
(0.07, 2.00) & 235.2 & 96.19 & 91.66 & 18.39 & 86.28 & 22.79 & 6.68 & 10.40 \\
(0.05, 1.00) & 238.7 & 95.15 & 90.42 & 19.51 & 82.26 & 30.87 & 3.06 & 9.33 \\
\bottomrule
\end{tabular}}
\end{table}

\section{Bridging Learning Curves: Ablation-Scale to Full-Scale}
\label{sec:supp_bridging}

\begin{figure*}[t]
\centering
\includegraphics[width=0.8\linewidth]{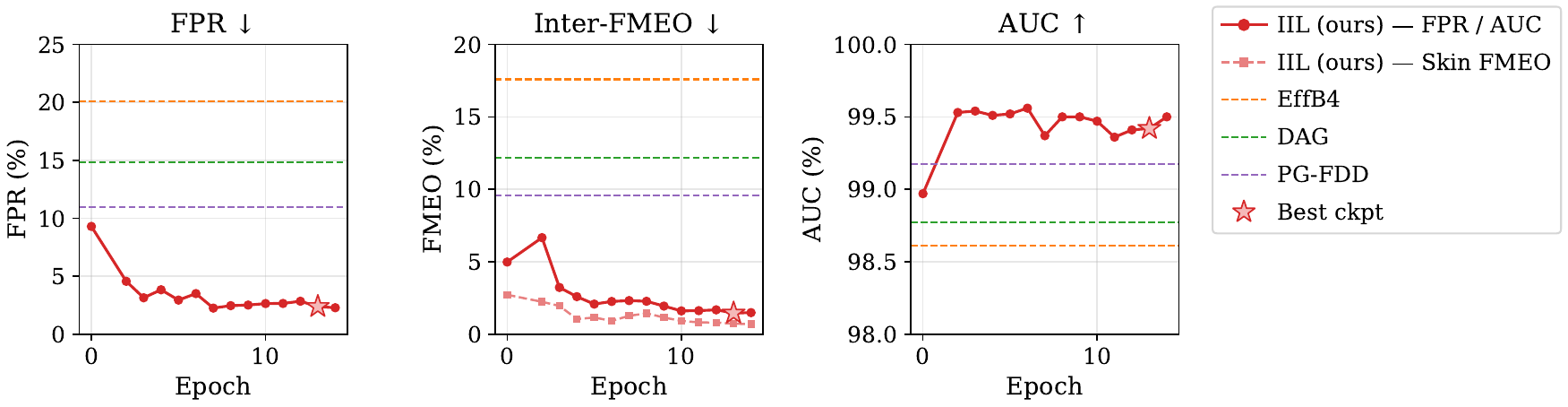}
\caption{IIL training curves over 15 epochs ($\lambda_\text{IRM}{=}0.07$,
$\lambda_\text{REx}{=}5.0$, $K{=}6$). Left: FPR. Centre: inter- and skin-FMEO.
Right: AUC. Star = best checkpoint (epoch~13, $S{=}6.09$).}
\label{fig:supp_training_curves}
\end{figure*}

Figure~\ref{fig:supp_training_curves} shows IIL's own training dynamics
over the full 15-epoch deployed run
($\lambda_\text{IRM}{=}0.07,\lambda_\text{REx}{=}5.0$, $K{=}6$): FPR drops
from 9.3\% to below 5\% by epoch~3, driven by $\mathcal{L}_{\text{FPR}}$,
which penalizes false positives on real images before IRM and REx have
reshaped the representation; inter-FMEO oscillates at epochs~4--7 as the
partition head reassigns groups while the encoder changes, settling by
epoch~9; AUC reaches 99.3\% by epoch~2 and stays there for the remainder
of training (98.97\% at epoch~0, still finishing its post-pretrain
warmup), confirming fairness and detection do not trade off once training
stabilizes.

\begin{figure*}[t]
\centering
\includegraphics[width=0.85\linewidth]{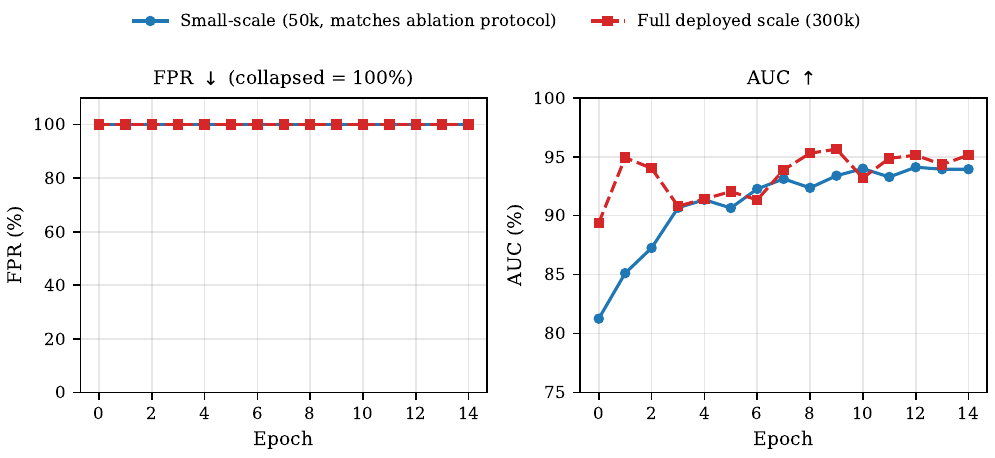}
\caption{Variant C (FPR+DP loss, no IRM/REx) at small (50k samples) and
full deployed (300k samples) scale, per-epoch over 15 epochs. Left: FPR
--- collapsed (100\%) at \emph{every} epoch, at both scales. Right: AUC
--- rises with training at both scales despite the FPR collapse, since
AUC measures ranking quality of raw scores, not the collapsed
classification threshold. Fairness score $S$ and FMEO are omitted: both
are trivially 0 under complete FPR collapse (uniform failure, not a
real parity result), matching Table~3's ``---'' convention for this
variant.}
\label{fig:supp_bridging}
\end{figure*}

The main text's fairness-utility result (17/21 metrics
won, including the \emph{highest} AUC of any compared method, main
paper \S5.2) is atypical for invariance methods and warrants a
mechanistic account beyond the result table alone.
\S5.4's explanation is that $\mathcal{L}_{\text{FPR}}$/$\mathcal{L}_{\text{DP}}$
directly optimize the measured fairness quantities while IRM/REx
stabilizes training rather than solely driving the gain;
Figure~\ref{fig:supp_bridging} connects the small-scale ablation
protocol (50k samples, 15 epochs) and the full deployed run (300k
samples, 15 epochs) on a shared FPR/AUC axis, to check whether variant C
(FPR+DP loss, no IRM/REx) still collapses at full data scale or whether
more data alone stabilizes it without invariance regularization.

\textbf{Result: variant C collapses at full deployed scale too, not
just small scale.} FPR is pinned at 100\% every single epoch at both
scales --- this is sustained collapse, not a transient early-training
artifact that later resolves. This directly supports \S5.4's
explanation: IRM/REx is not compensating for insufficient data or short
training; without it, the parity losses overwhelm the classification
objective regardless of scale. More data alone does not stabilize
$\mathcal{L}_{\text{FPR}}$/$\mathcal{L}_{\text{DP}}$ without IRM/REx's
invariance penalty.

\section{Age-Bracket Composition Within the Six Optimized Partitions}
\label{sec:supp_age_composition}

IIL's FDP worsens on Skin~Tone, Age, and Intersection (
Table~2) despite our losses directly targeting FPR/DP parity; this
section gives a non-post-hoc, structural account of why. \S5.2 argues this is a
structural consequence of the partition choice: $\mathcal{L}_{\text{FPR}}$
and $\mathcal{L}_{\text{DP}}$ equalize outcomes across the six
gender$\times$skin-tone partitions used for training, but Age is not
nested within those six partitions, so a partition-level constraint does
not imply an Age-bracket-level one. Table~\ref{tab:supp_age_composition}
verifies this is a real property of the training distribution, not just
a plausible-sounding mechanism: the within-partition Age-bracket mix
varies substantially across the six partitions.

\begin{table}[h]
\centering
\footnotesize
\caption{Row-normalized Age-bracket composition (\%) within each of the
six gender$\times$skin-tone partitions used by $\mathcal{L}_{\text{FPR}}$/
$\mathcal{L}_{\text{DP}}$/IRM. Each row sums to 100\%.}
\label{tab:supp_age_composition}
\begin{tabular}{>{\raggedright\arraybackslash}p{1.2cm}*{5}{>{\centering\arraybackslash}p{0.85cm}}}
\toprule
Partition & Child & Young & Adult & Middle & Senior \\
\midrule
F, skin~1 & 9.18 & 35.11 & 42.61 & 10.66 & 2.44 \\
F, skin~2 & 4.98 & 25.69 & 48.78 & 18.38 & 2.18 \\
F, skin~3 & 4.58 & 20.01 & 53.06 & 18.74 & 3.60 \\
M, skin~1 & 8.34 & 20.25 & 44.94 & 20.45 & 6.02 \\
M, skin~2 & 2.38 & 10.08 & 49.08 & 33.76 & 4.70 \\
M, skin~3 & 1.28 & 8.18 & 50.92 & 35.52 & 4.10 \\
\midrule
range & 1.28--9.18 & 8.18--35.11 & 42.61--53.06 & 10.66--35.52 & 2.18--6.02 \\
\bottomrule
\end{tabular}
\end{table}

Every Age bracket's share swings several-fold across the six partitions
(e.g.\ Child: 1.3\%--9.2\%; Middle: 10.7\%--35.5\%; Young: 8.2\%--35.1\%).
Concretely, F/skin-1 skews young (Child+Young $=$ 44.3\% of that
partition, Senior only 2.4\%) while M/skin-3 skews toward Middle-aged
(Child+Young $=$ 9.5\%, Middle alone $=$ 35.5\%). Enforcing FPR/DP parity
\emph{across the six rows} of this table is compatible with arbitrarily
unequal FPR/DP \emph{across Age brackets within a row}, since the
optimization never observes the Age column. This is a property of the
partition definition intersecting the dataset's demographic
distribution, established before training, not an artifact fit to the
observed FDP gap after the fact.

\section{Quantitative Representation-Invariance Probe}
\label{sec:supp_invariance_probe}

The UMAP/GradCAM evidence in \S5.5 is qualitative; this section adds a
quantitative analysis of representation invariance via a
demographic-group classifiability probe: a linear classifier (5-fold
cross-validated logistic regression on a 50-component PCA of each
model's penultimate-layer features, which puts the four models'
differently-sized native feature spaces, 512--2048 dimensions, on equal
footing) trained to predict demographic group directly from the learned
features. Lower accuracy means group is harder to recover from the
representation, i.e.\ more invariant in this specific sense. Evaluated
on the same 4,713-image sample used for the paper's UMAP figure
(\S5.5), balanced across the six groups.

\begin{table}[h]
\centering
\footnotesize
\caption{Demographic-group classifiability from learned features
(accuracy, mean$\pm$std over 5 folds; lower $=$ more invariant).
Majority-class baseline given for reference.}
\label{tab:supp_invariance_probe}
\begin{tabular}{>{\raggedright\arraybackslash}p{1.3cm}*{3}{>{\centering\arraybackslash}p{1.3cm}}}
\toprule
Model & Intersection (6-way) & Skin (3-way) & Gender (2-way) \\
\midrule
Majority-class baseline & 17.0\% & 33.9\% & 50.7\% \\
\midrule
EffB4      & 55.1$\pm$1.0\% & 75.6$\pm$0.7\% & 71.7$\pm$1.3\% \\
DAG-FDD    & 37.7$\pm$1.1\% & 62.9$\pm$1.2\% & 61.5$\pm$2.0\% \\
PG-FDD     & \textbf{22.1$\pm$1.2\%} & \textbf{41.9$\pm$1.0\%} & \textbf{54.4$\pm$1.5\%} \\
IIL (ours) & 41.1$\pm$0.6\% & 60.1$\pm$0.8\% & 66.3$\pm$0.9\% \\
\bottomrule
\end{tabular}
\end{table}

\textbf{IIL is not the most invariant representation by this metric.}
PG-FDD's features are substantially harder to decode group from on all
three targets; IIL is closer to DAG-FDD than to PG-FDD, though clearly
better than the unconstrained EffB4 baseline. This does not contradict
Table~2, where IIL wins FMEO/FPR/DP decisively, because
IRM/REx (Eqs.~1--2, \S4.1) and this probe measure
different properties. IRM/REx constrain the classifier's
\emph{conditional risk} to be stable across partitions --- the gradient
penalty (Eq.~1) operates only on the scalar output, never on
$\Phi(x)$ directly --- so a classifier can become invariant to which
partition-specific cue it \emph{exploits} without demographic identity
becoming any less recoverable from the representation by a different
probe. This is the same distinction the original IRM paper illustrates
with colored MNIST: the model stops \emph{using} color to predict digit
identity, but color remains fully decodable from the representation.
Methods built around explicit representation alignment (plausibly closer
to PG-FDD's mechanism, though we have not verified this against its
published method) target exactly the quantity this probe measures;
IIL's mechanism never did, so this result is consistent with, not a
failure of, IIL's actual design. We report it because it is the
correct quantitative complement to the paper's UMAP figure's
qualitative picture, not because it favors IIL.

\end{document}